\documentclass[11pt,a4paper]{article}
\usepackage{authblk} 

\usepackage[hyperref]{naaclhlt2019}
\usepackage{times}
\usepackage{latexsym}
\usepackage{xcolor}
\usepackage{color}
\usepackage{url}

\aclfinalcopy 

\usepackage{graphicx}
\usepackage{tabularx}
\usepackage{booktabs}
\usepackage{url}
\usepackage{dsfont}
\usepackage{todonotes}
\usepackage{amsmath,amsfonts,amssymb}
\usepackage{subfigure}

\definecolor{cerulean}{rgb}{0.0, 0.48, 0.65}
\definecolor{ceruleanblue}{rgb}{0.16, 0.32, 0.75}
\definecolor{carnelian}{rgb}{0.7, 0.11, 0.11}

\title{Learning Graph Embeddings from WordNet-based Similarity Measures}

\author[1]{\textbf{Andrey Kutuzov}}
\author[2]{\textbf{Mohammad Dorgham}}
\author[2]{\textbf{Oleksiy Oliynyk}}
\author[2]{\\\textbf{Chris Biemann}}
\author[2,3]{\textbf{Alexander Panchenko}}
\affil[1]{Language Technology Group, University of Oslo, Oslo, Norway}
\affil[2]{Language Technology Group, University of Hamburg, Hamburg, Germany}
\affil[3]{Skolkovo Institute of Science and Technology, Moscow, Russia}

 \affil[ ]{\href{mailto:andreku@ifi.uio.no}{andreku@ifi.uio.no}}
 \affil[ ]{\href{mailto:panchenko@informatik.uni-hamburg.de}{\{5dorgham,6oliinyk,biemann,panchenko\}@informatik.uni-hamburg.de}}

\begin{document}
\maketitle
\begin{abstract}
We present \textit{path2vec}, a new approach for learning graph embeddings that relies on structural measures of pairwise node similarities. The model learns representations for nodes in a dense space that approximate a given user-defined graph distance measure, such as e.g. the shortest path distance or distance measures that take information beyond the graph structure into account. Evaluation of the proposed model on semantic similarity and word sense disambiguation tasks, using various WordNet-based similarity measures, show that our approach yields competitive results, outperforming strong graph embedding baselines. The model is computationally efficient, being orders of magnitude faster than the direct computation of graph-based distances.
\end{abstract}

\section{Introduction }

Developing applications making use of large graphs, such as networks of roads, social media users, or word senses, often involves the design of a domain-specific graph \textbf{node similarity measure} $sim: V\times V \rightarrow \mathds{R}$ defined on a set of nodes $V$ of a graph $G = (V, E)$. For instance, it can represent the shortest distance from home to work, a community of interest in a social network for a user, or a semantically related sense to a given synset in WordNet \cite{wordnet:1993}. There exist a wide variety of such measures greatly ranging in their complexity and design from simple deterministic ones, e.g. based on shortest paths in a network~\cite{leacock1998combining} to more complex ones, e.g. based on random walks~\cite{fouss2007random,pilehvar2015senses,lebichot2018constrained}. Naturally, the majority of such measures rely on walks along edges $E$ of the graph, often resulting in effective, but prohibitively inefficient measures requiring complex and computationally expensive graph traversals. Also, there are measures that in addition take e.g. corpus information into account beyond what is directly given in the graph, see e.g. \cite{hirst2006}. We propose a solution to this problem by decoupling development and use of graph-based measures. Namely, once a node similarity measure is defined, we learn vector representations of nodes that enable efficient computation of this measure.     
We represent nodes in a graph with dense embeddings that are good in approximating such custom, e.g. application-specific, pairwise node similarity measures. 
Similarity computations in a vector space are several orders of magnitude faster than computations directly using the graph.  Additionally, graph embeddings can be of importance in privacy-sensitive network datasets, since in this setup, explicitly storing edges is not required anymore. 
The main advantage over other graph embeddings is that our model can learn a custom user-defined application or domain specific similarity measure.

We show the effectiveness of the proposed approach \textit{intrinsically} on a word similarity task, by learning synset vectors of the WordNet graph based on several similarity measures. Our model is not only able to closely \textit{approximate} various measures, but also \textit{to improve} the results of the original measures in terms of (1) correlation with human judgments and (2) computational efficiency, with gains up to 4 orders of magnitude. Our method outperforms other state-of-the-art graph embeddings models. 

Besides, we evaluate it \textit{extrinsically} in a WSD task~\cite{navigli2009word} by replacing the original structural measures with their vectorized counterparts in a graph-based WSD algorithm by \citet{sinha2007unsupervised}, reaching comparable performance. Because of being inspired by the \textit{word2vec} architecture, we dub our model `\textit{path2vec}'\footnote{\url{https://github.com/uhh-lt/path2vec}} meaning it encodes paths (or other similarities) between graph nodes into dense vectors.

Our \textbf{first contribution} is an effective and efficient approach to learn graph embeddings based on a user-defined custom similarity measure $sim$ on a set of nodes $V$, e.g. the shortest path distance. The \textbf{second contribution} is an application of state-of-the-art graph embeddings to word sense disambiguation task. 

\section{Related Work} \label{sec:related}
Various methods have been employed in NLP to derive lexical similarity directly from geometrical properties of the WordNet graph, from random walks in \cite{affinity:2008} to kernels in \cite{kernels:2009}. More recently, representation learning on graphs \cite{bordes2011learning} received much attention in various research communities; see \cite{hamilton2017representation} for a thorough survey on the existing methods. All of them (including ours) are based on the idea of projecting graph nodes into a latent vector space with a much lower dimensionality than the number of nodes.

The method described in this paper falls into the category of `shallow embeddings', meaning that we do not attempt to embed entire communities or neighborhoods: our aim is to approximate distances or similarities between (single) nodes. Existing approaches to solving this task mostly use either factorization of the graph adjacency matrix  \cite{cao2015grarep,ou2016asymmetric} or random walks over the graph as in \textit{Deepwalk} \cite{perozzi2014deepwalk} and \textit{node2vec} \cite{grover2016node2vec}. A completely different approach is taken by \citet{subercaze:2015}, who directly embed the WordNet tree graph into Hamming hypercube binary representations. Their model is dubbed `Fast similarity embedding' (\textit{FSE}) and also optimizes one of our objectives, i.e. to provide a much quicker way of calculating semantic similarities based on WordNet knowledge. However, the \textit{FSE} embeddings are not differentiable, limiting their use in many deep neural architectures, especially if fine-tuning is needed. 

TransE \cite{transe:2013} interprets entities as vectors in the low-dimensional embeddings space and relations as a translation operation between two entity vectors. For a triplet (head, relation, tail) which holds, the embedding of the tail is close to the embedding of the head plus embedding of the relation. 
TransH \cite{transh:2014}  models each relation as a specific hyperplane and projects entity vectors onto the hyperplane. If connection holds then projected vectors of head and tail are connected by a translation vector with low error. As a result, entities have different representations for hyperplanes of different relations where they are involved. \textit{TransR} \cite{transr:2015} extends \textit{TransE} \cite{transe:2013} and \textit{TransH} \cite{transh:2014}, and is based on the idea that an entity may have a few aspects and different relations are focused on them. So the same entities can be close or far from each other depending on the type of the relation. 
\textit{TransR} projects entity vectors into a relation specific space, and learns embeddings via translation between projected entities.

We quantitatively compare \textit{path2vec} to these methods in Section \ref{sec:experimental}. We did not compare our approach to the \textit{GraphSAGE} embeddings \cite{hamilton2017inductive} and Graph Convolutional Networks \cite{kipf2018modeling}, since they make use of input node features, which are absent in our setup.

Also note that unlike retro-fitting and similar techniques \cite{autoextend:2015,deconf:2016,retrofitting:2017}, our approach does not use any training corpus or pre-trained input embeddings. The synset representations are trained on the WordNet graph alone.

\section{Learning Graph Metric Embeddings}
\label{subsec:model}

\paragraph{Definition of the Model} 

The \textit{path2vec} model learns low-dimensional vectors for the graph nodes $\{v_i,v_j\} \in V$ (synsets in the case of WordNet) such that the dot products between pairs of the respective vectors $(\mathbf{v}_i \cdot \mathbf{v}_j)$ are close to the user-defined similarities between the nodes $s_{ij}$. This first component of the objective encodes potentially long distances in the graph (the \textcolor{carnelian}{{global structure}}). In addition, the model reinforces direct connections between nodes: We add to the objective similarities $\mathbf{v}_i \cdot \mathbf{v}_{n}$ and $ \mathbf{v}_j \cdot \mathbf{v}_{m}$ between the nodes $v_i$ and $v_j$ and their respective adjacent nodes $\{v_{n}: \exists (v_i, v_{n})\in E\}$ and $\{v_{m}: \exists (v_j, v_{m})\in E\}$ to preserve \textcolor{ceruleanblue}{{local structure}} of the graph. Therefore, the model preserves both \textcolor{carnelian}{{global}} and \textcolor{ceruleanblue}{{local}} relations between nodes by minimizing the following loss function $\mathcal{L}$:

\vspace{-10pt}
$$
\frac{1}{|B|}  \sum_{ (v_i, v_j) \in B }\left((\textcolor{carnelian}{\mathbf{v}_i^T \mathbf{v}_j - s_{ij}} )^2 - \alpha (\textcolor{ceruleanblue}{\mathbf{v}_i^T \mathbf{v}_{n}} +  \textcolor{ceruleanblue}{\mathbf{v}_j^T \mathbf{v}_{m}})\right),
$$

where $s_{ij} = sim(v_i, v_j)$ is the value of a `gold' similarity measure between a pair of nodes $v_i$ and $v_j$,  $\mathbf{v}_i$ and  $\mathbf{v}_j$ are the embeddings of the first and the second node, $B$ is a training batch, 
$\alpha$ is a regularization coefficient. The second term ($\textcolor{ceruleanblue}{\mathbf{v}_i \cdot \mathbf{v}_{n}} +  \textcolor{ceruleanblue}{\mathbf{v}_j \cdot \mathbf{v}_{m}}$) in the objective function is a regularizer which aids the model to simultaneously maximize the similarity between adjacent nodes (which is maximum by definition) while learning the similarity between the two target nodes.

We use negative sampling to form a training batch $B$  adding $n$ negative samples ($s_{ij}=0$) for each  real ($s_{ij} > 0$) training instance: each real node (synset) pair $(v_i, v_j)$ with `gold' similarity $s_{ij}$ is accompanied with $n$ `negative' node pairs $(v_i, v_k)$ and $(v_j, v_l)$ with zero similarities, where $v_k$ and $v_l$ are randomly sampled nodes from $V$. Embeddings are initialized randomly and trained with the  \textit{Adam} optimizer \cite{kingma2014adam} with early stopping.\footnote{In our experiments, we identified the optimal values of $n=3$ negative samples, batch size of $|B|=100$, training for 15 epochs, $\alpha=0.01$. We report on the influence of the embedding dimensionality parameter $d$ in Section~\ref{sec:results}. We found it also beneficial to use additionally $L_1$ weight regularization. } 

Once the model is trained, the computation of node similarities is approximated with the dot product of the learned node vectors, making the computations efficient: $\hat{s}_{ij} = \mathbf{v}_i \cdot \mathbf{v}_j$.

\paragraph{Relation to Similar Models} Our model is similar to the Skip-gram model \cite{Mikolov:2013}, where pairs of words $(v_i, v_j)$ from a training corpus are optimized to have their corresponding vectors dot product $\mathbf{v}_i \cdot  \mathbf{\Tilde{v}}_j$ close to 1, while randomly generated pairs (`negative samples') are optimized to have their dot product close to 0.  In the Skip-gram model, the target is to minimize the log likelihood of the conditional probabilities of context words $v_j$ given current words $v_i$, which is in the case on the negative sampling amounts to minimization of: $\mathcal{L} = - \sum_{ (v_i, v_j) \in B_p } \log \sigma{( \textcolor{ceruleanblue}{\mathbf{v}_i \cdot \mathbf{\Tilde{v}}_j}  )  } - \sum_{ (v_i, v_j) \in B_n } \log \sigma{(  \textcolor{ceruleanblue}{ \mathbf{-v}_i \cdot \mathbf{\Tilde{v}}_j}   )}\Big)$,
where $B_p$ is the batch of positive training samples, $B_n$ is the batch of the generated negative samples, and $\sigma$ is the sigmoid function. The model uses \textcolor{ceruleanblue}{local} information. However, in \textit{path2vec}, the target values $s_{ij}$ for the dot product are not binary, but can take arbitrary values in the $[0...1]$ range, depending on the path-based measure on the input graph, e.g. the normalized shortest path length in the WordNet between  \textit{motor.n.01}  and \textit{rocket.n.02} is 0.389.

Further, in our model there is no difference between `word' and `context' spaces: we use a single embedding matrix, with the number of rows equal to the number of nodes and column width set to the desired embedding dimensionality. Finally, unlike the Skip-gram, we do not use any non-linearities.

Another closely related model is Global Vectors (GloVe) \cite{pennington2014glove}, which approximates the co-occurrence probabilities in a given corpus. The objective function to be minimized in GloVe model is $\mathcal{L} = \sum_{ (v_i, v_j) \in B } f( \textcolor{carnelian}{ s_{ij} } ) (  \textcolor{carnelian}{ {\mathbf{v}_i \cdot \mathbf{\Tilde{v}}_j - \log s_{ij} } } + b_i + b_j  )^2$, where $s_{ij}$ counts the number of co-occurrence of words $v_i$ and $v_j$, $b_i$ and $b_j$ are additional biases for each word, and $f(s_{ij})$ is a weighting function to give appropriate weight for rare co-occurrences. Like the Skip-gram, GloVe also uses two embedding matrices, but it relies on \textcolor{carnelian}{global} information.

\section{Computing Pairwise Similarities}
\label{sec:datasets}

\subsection{Selection of the Similarity Measures}
Our aim is to produce node embeddings that capture given similarities between nodes in a graph. In our case, the graph is WordNet, and the nodes are its 82,115 noun synsets. We focused on nouns since in WordNet and SimLex999 they are represented better than other parts of speech. Embeddings for synsets of different part of speech can be generated analogously.
    
The training datasets consist of pairs of noun synsets and their `ground truth' similarity values.  There exist several methods to calculate synset similarities on the WordNet \cite{hirst2006}. We compile four datasets, with different similarity functions: Leacock-Chodorow similarities (\textit{LCH});  Jiang-Conrath similarities calculated over the SemCor corpus (  \textit{JCN-S});  Wu-Palmer similarities (\textit{WuP}); and  Shortest path similarities (\textit{ShP}). \textit{LCH} similarity \cite{leacock1998combining} is based on the shortest path between two synsets in the WordNet hypernym/hyponym taxonomy and its maximum depth, while \textit{JCN} similarity \cite{jiang1997} uses the lowest common parent of two synsets in the same taxonomy. \textit{JCN} is significantly faster but additionally requires a corpus as a source of probabilistic data about the distributions of synsets (`information content'). We employed the SemCor subset of the Brown corpus, manually annotated with word senses \cite{kucera1982frequency}.

\textit{WuP} similarities \cite{wupalmer:1994} are based on the depth of the two nodes in the taxonomy and the depth of their most specific ancestor node. \textit{ShP} is a simple length of the shortest path between two nodes in the graph. We used the NLTK \cite{bird2009natural} implementations of all the aforementioned similarity functions. 

Pairwise similarities for all synset pairs can be pre-computed. For the 82,115 noun synsets in the WordNet, this results in about 3 billion unique synset pairs. Producing these similarities using 10 threads takes about 30 hours on an Intel Xeon E5-2603v4@1.70GHz CPU for \textit{LCH}, and about 5 hours for \textit{JCN-S}. The resulting similarities lists are quite large (45 GB compressed each) and thus difficult to use in applications. 
But they can be used in \textit{path2vec} to \textit{learn} dense embeddings $\mathbb{R}^d$ for these  82,115 synsets, such that $d \ll 82,115$ and the dot products between the embeddings approximate the `raw' WordNet similarity functions. 

\subsection{Pruning the Dissimilar Pairs of Nodes}
\begin{figure}
       \centering
       \includegraphics[scale=0.5,keepaspectratio]{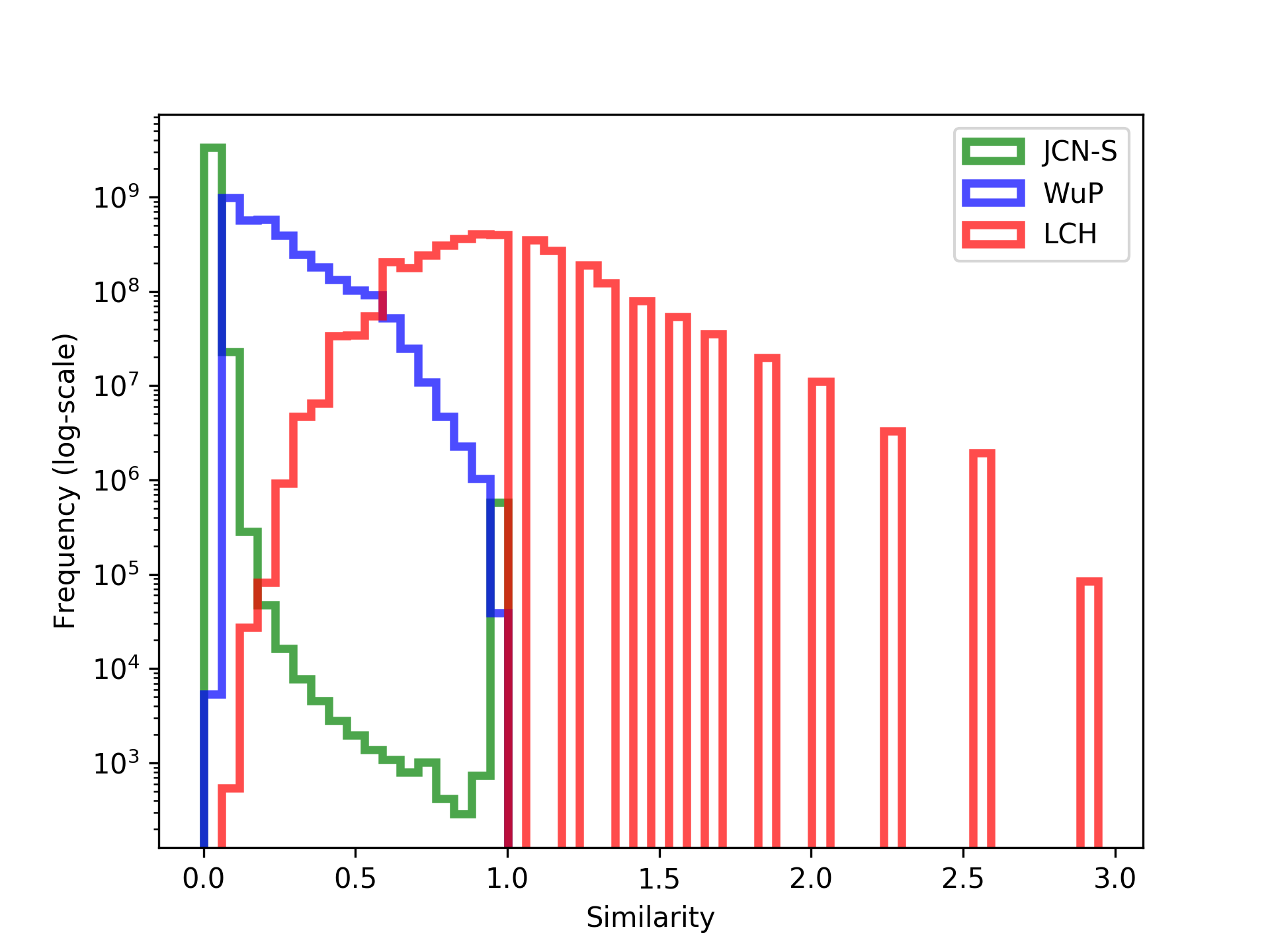}
       \caption{Distribution of similarities between WordNet noun synsets with different distance measures.}
       \label{fig:histo}
 \end{figure}

In principle, one can use all unique synset pairs with their WordNet similarities as the training data. However, this seems impractical. As expected due to the small-world nature of the WordNet graph \cite{Steyvers2005}, most synsets are not similar at all: with \textit{JCN-S}, the overwhelming majority of pairs feature similarity very close to zero; with \textit{LCH}, most pairs have similarity below 1.0. Thus, we filtered low-similarity pairs out, using similarity threshold of 0.1 for the \textit{JCN-S} and \textit{ShP} datasets, 0.3 for the \textit{WuP} dataset and 1.5 for the \textit{LCH} dataset (due to substantial differences in similarities distributions, as shown in Figure \ref{fig:histo}). This dramatically reduced the size of the training data (e.g., to less than 1.5 million pairs for the \textit{JCN-S} dataset and to 125 million pairs for the \textit{LCH} dataset), thus making the training much faster and at the same time \textit{improving} the quality of the resulting embeddings (see the description of our evaluation setup below). 

With this being the case, we additionally pruned these reduced datasets by keeping only the first 50 most similar `neighbors' of each synset:  the rationale behind this is that some nodes in the WordNet graph are very central and thus have many neighbors with high similarity, but for our procedure only the nearest/most similar ones suffice. This again reduced training time and improved the results, so we hypothesize that such pruning makes the models more generally applicable and more focused on the meaningful relations between synsets. The final sizes of the pruned training datasets are 694,762 pairs for the \textit{JCN-S}, 4,008,446 pairs for the \textit{LCH}, 4,063,293 pairs for the \textit{ShP} and 4,100,599 pairs for the \textit{WuP}\footnote{All the datasets and the trained graph embeddings can be downloaded from \url{https://github.com/uhh-lt/path2vec}}.

\begin{figure}
       \centering
       \includegraphics[scale=0.5,keepaspectratio]{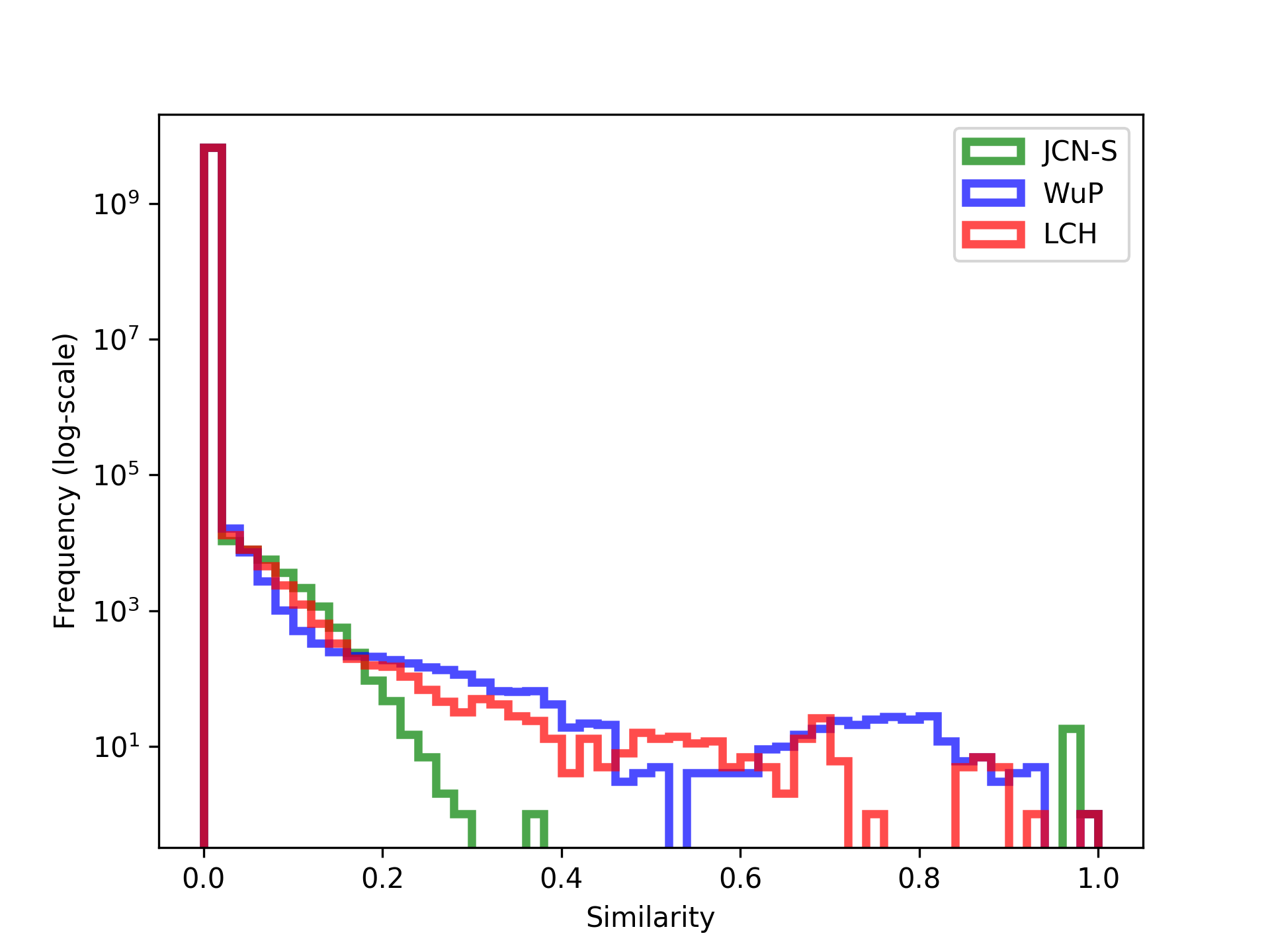}
       \caption{Distributions of pairwise similarities in \textit{path2vec} models trained on different measures.}
       \label{fig:p_histo}
 \end{figure}

Note also that the \textit{LCH} similarity can take values well above 1.0. 
After the pruning, we scaled similarities in all datasets to the $[0...1]$ range by unity-based normalization. Also, in some rare cases, NLTK produces \textit{JCN} similarities of infinitely large values (probably due to the absence of particular synsets in SemCor). We clipped these similarities to the value of 1. All the datasets were shuffled prior to training. 

\section{Experiment 1: Intrinsic Evaluation based on Semantic Similarity}
\label{sec:experimental}

\paragraph{Experimental Setting}
It is possible to evaluate the models by calculating the rank correlation of their cosine similarities with the corresponding similarities for all the unique pairs from the training dataset, or at least a large part of them. \newcite{subercaze:2015} evaluated their approach on \textit{LCH} similarities for all unique noun synset pairs from WordNet Core (about 5 million similarities total); their model achieves Spearman rank correlation of 0.732 on this task. However, this kind of evaluation does not measure the ability of the model to produce meaningful predictions, at least for language data: the overwhelming part of these unique pairs are synsets not related to each other at all. For most tasks, it is useless to `know' that, e.g., `\textit{ambulance}' and `\textit{general}' are less similar than `\textit{ambulance}' and `\textit{president}'. While the distances between these node pairs are indeed different on the WordNet graph, we find it much more important for the model to be able to robustly tell really similar pairs from the unrelated ones so that they could benefit applications.

As a more balanced and relevant test set, we use noun pairs (666 total) from the SimLex999 semantic similarity dataset \cite{Hill2015}. SimLex999 contains lemmas; as some lemmas may map to several WordNet synsets, for each word pair we choose the synset pair maximizing the WordNet similarity, following \cite{Resnik1999}. Then, we measure the Spearman rank correlation between these `gold' scores and the similarities produced by the graph embedding models trained on the WordNet. Further on, we call this evaluation score the `correlation with WordNet similarities'. This evaluation method directly measures how well the model fits the training objective\footnote{Note, however, that it does not mean testing on the training data: for example, 75\% of synset pairs from the SimLex999 are \textbf{not present} in our pruned \textit{JCN-S} training dataset; for the \textit{LCH} dataset it is 82\%. Evaluating these absent pairs only does not substantially change the results.}. 

We also would like to check whether our models generalize to extrinsic tasks. Thus, we additionally used human-annotated semantic similarities from the same SimLex999. This additional evaluation strategy directly tests the models' correspondence to human judgments independently of WordNet. These correlations were tested in two synset selection setups, important to distinguish:

\textbf{1. WordNet-based synset selection} (static synsets): this setup uses the same lemma-to-synset mappings, based on maximizing WordNet similarity for each SimLex999 word pair with the corresponding similarity function. It means that all the models are tested on exactly the same set of synset pairs (but the similarities themselves are taken from SimLex999, not from the WordNet).   
 
\textbf{2. Model-based synset selection} (dynamic synsets): in this setup, lemmas are converted to synsets dynamically as a part of the evaluation workflow. We choose the synsets that maximize word pair similarity \textit{using the vectors from the model itself}, not similarity functions on the WordNet. Then the resulting ranking is evaluated against the original SimLex999 ranking.

The second (dynamic) setup in principle allows the models to find better lemma-to-synset mappings than those provided by the WordNet similarity functions. This setup essentially evaluates two abilities of the model: 1) to find the best pair of synsets for a given pair of lemmas (sort of a disambiguation task), and 2) to produce the similarity score for the chosen synsets. We are not aware of any `gold' lemma-to-synset mapping for SimLex999, thus we directly evaluate only the second part, but implicitly the first one still influences the resulting scores. Models often choose different synsets. For example, for the word pair `\textit{atom, carbon}', the synset pair maximizing the \textit{JCN-S} similarity calculated on the `raw' WordNet would be `atom.n.02 (\textit{`a tiny piece of anything'}), carbon.n.01 (\textit{`an abundant nonmetallic tetravalent element'})' with the similarity 0.11. However, in a \textit{path2vec} model trained on the same gold similarities, the synset pair with the highest similarity 0.14 has a different first element: `atom.n.01 (\textit{`the smallest component of an element having the chemical properties of the element'})', which seems to be at least as good a decision as the one from the raw WordNet.
 
\paragraph{Baselines}
\textit{path2vec} is compared against five baselines (more on them in Section \ref{sec:related}): {\em raw WordNet similarities} by respective measures;
 {\em Deepwalk} \cite{perozzi2014deepwalk}; {\em node2vec} \cite{grover2016node2vec}; \textit{FSE} \cite{subercaze:2015}; and
{\em TransR} \cite{transr:2015}.

\textit{DeepWalk}, \textit{node2vec}, and \textit{TransR} models were trained on the same WordNet graph. We used all 82,115 noun synsets as vertices and hypernym/hyponym relations between them as edges. Since the \textit{node2vec} C++ implementation accepts an edge list as input, we had to add a self-connection for all nodes (synsets) that lack edges in WordNet. During the training of \textit{DeepWalk} and \textit{node2vec} models, we tested different values for the number of random walks (in the range from 10 to 100), and the vector size (100 to 600). For \textit{DeepWalk}, we additionally experimented with the window size (5 to 100). All other hyperparameters were left at their default values. \textit{FSE} embeddings of  the WordNet noun synsets were provided to us by the authors, and consist of 128-bit vectors.

\begin{table}

\footnotesize
\centering 
\begin{tabular}{lcccc}
\toprule
& \multicolumn{4}{c}{\textit{Selection of synsets}} \\
Model & JCN-S & LCH & ShP & WuP \\
\midrule
WordNet & 1.0  & 1.0 & 1.0 & 1.0  \\
\midrule
TransR & 0.568  & 0.776 & 0.776 & 0.725 \\
node2vec & 0.726 & 0.759 & 0.759 & 0.787  \\
Deepwalk & 0.775 & 0.868 & 0.868 & 0.850 \\
FSE & 0.830  & 0.900 & 0.900 & 0.890  \\
\midrule
path2vec & \textbf{0.931} & \textbf{0.935} & \textbf{0.952} & \textbf{0.931}  \\
\bottomrule
\end{tabular}
\caption {Spearman correlation scores with WordNet similarities on the 666 noun pairs in SimLex999.} 
\label{tab:eval_wordnet}
\end{table}

\paragraph{Discussion of Results}
\label{sec:results}

Table \ref{tab:eval_wordnet} presents the comparison of \textit{path2vec} and the baselines with regards to how well they approximate the WordNet similarity functions output (the raw WordNet similarities always get the perfect correlation in this evaluation setup). All the reported rank correlation value differences in this and other tables are statistically significant based on the standard two-sided \textit{p}-value. We report the results for the best models for each method, all of them (except \textit{FSE}) using vector size 300 for comparability. 

\begin{figure*}[ht]
    \centering
       \subfigure{\includegraphics[scale=0.487,keepaspectratio]{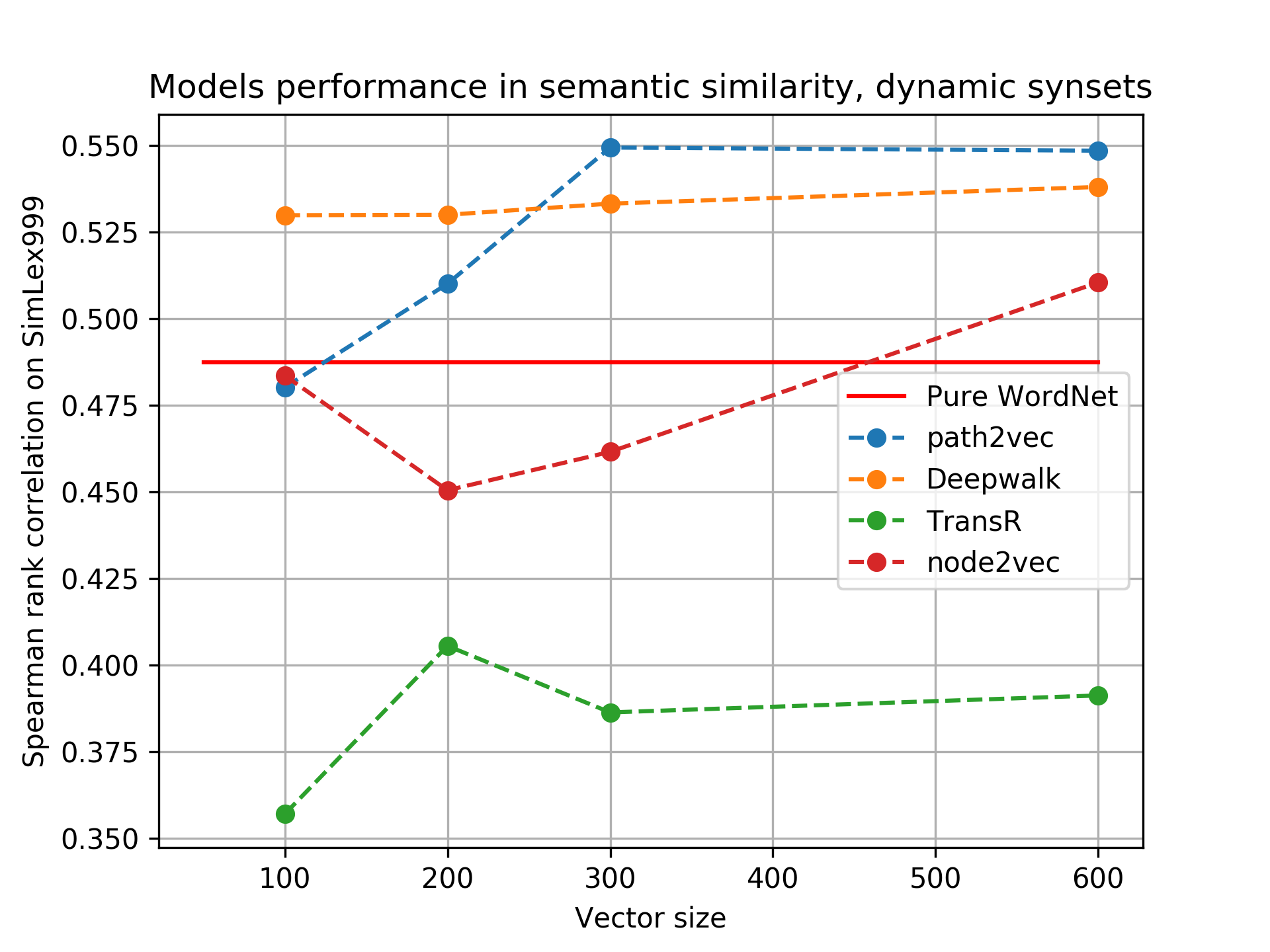}}
       \subfigure{\includegraphics[scale=0.487,keepaspectratio]{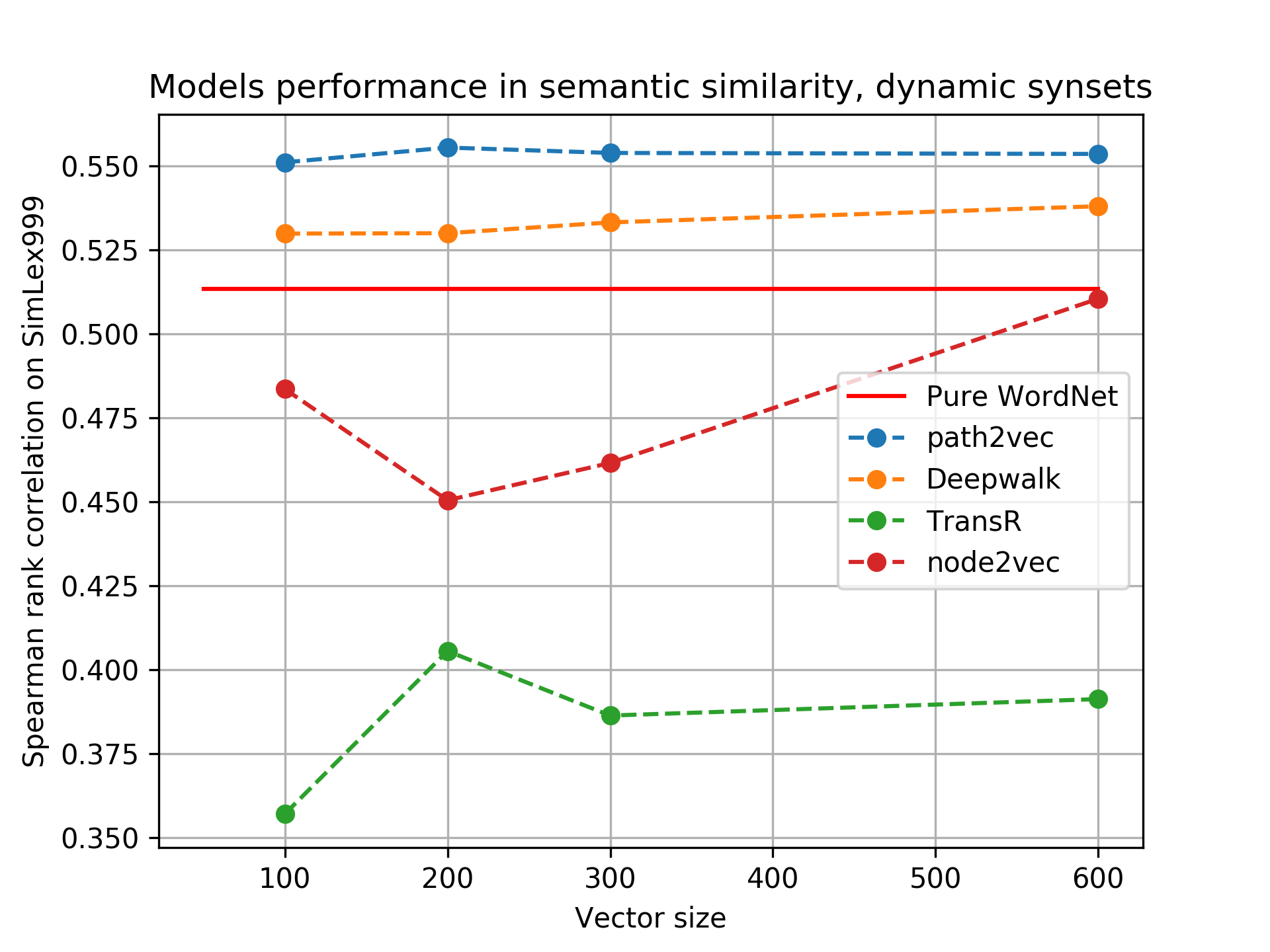}}
       
\vspace{-5pt}
       \caption{Evaluation on SimLex999 noun pairs, model-based synset selection: \textit{JCN-S} (left) and \textit{WuP} (right).}
       \label{fig:dynamic}
\end{figure*}

\textit{Path2vec} outperform other baseline embeddings, achieving high correlation with the raw WordNet similarities. This shows that our  simple model can  approximate  different graph measures. Figure \ref{fig:p_histo} shows the similarities' distributions in the resulting models, reflecting the original measures' distributions in Figure \ref{fig:histo}.

As expected, vector dimensionality greatly influences the performance of all graph embedding models. As an example, Figure \ref{fig:dynamic} plots the performance of the \textit{path2vec} models trained on \textit{JCN-S} and \textit{WuP} datasets, when using `dynamic synset selection' evaluation setup (that is, each model can decide for itself how to map SimLex999 lemmas to WordNet synsets). The red horizontal line is the correlation of WordNet similarities with SimLex999 human scores. For the \textit{path2vec} models, there is a tendency to improve the performance when the vector size is increased, until a plateau is reached beyond 600 dimensions.  Note that \textit{Deepwalk}\footnote{The reported best \textit{Deepwalk} models were trained with the number of walks 10 and window size 70.} does not benefit much from increased vector size, while \textit{node2vec}\footnote{The reported best \textit{node2vec} models were trained with the number of walks 25.} yields strangely low scores for 200 dimensions. Interestingly, \textit{path2vec} and \textit{Deepwalk} models consistently \textit{outperform} the raw WordNet (this is also true for \textit{FSE}). This means these embeddings are in some sense `regularized', leading to better `disambiguation' of senses behind SimLex999 word pairs and eventually to better similarities ranking. 

\begin{table}
\footnotesize
\centering
    \begin{tabular}{lcccc}
    \toprule
    & \multicolumn{4}{c}{\textit{Selection of synsets}} \\
    Model & JCN-S & LCH & ShP & WuP \\
    \midrule
    WordNet & 0.487 &  \textbf{0.513} & \textbf{0.513} & 0.474\\
    \midrule
    TransR & 0.394 & 0.395 & 0.395 & 0.379 \\
    node2vec & 0.426 & 0.434 & 0.434 & 0.400 \\
    Deepwalk & 0.468 & 0.468 & 0.468 & 0.450 \\
    FSE & 0.490 & 0.502 & 0.502 & 0.483 \\
    \midrule
    path2vec & \textbf{0.501} & 0.470 & 0.512 & \textbf{0.491} \\
    \bottomrule
    \end{tabular}
\caption {Spearman correlations with human SimLex999 noun similarities (WordNet synset selection).}
\label{tab:eval_wordnet_based}
\end{table}

In Tables \ref{tab:eval_wordnet_based} and \ref{tab:eval_model_based}, we select the best 300D \textit{path2vec} models from the experiments described above and compare them against the best 300D baseline models and 128D FSE embeddings in static and dynamic evaluation setups. When WordNet-defined lemma-to-synset mappings are used (Table \ref{tab:eval_wordnet_based}), the raw WordNet similarities are non-surprisingly the best, although \textit{FSE} and \textit{path2vec} embeddings achieve nearly the same performance (even slightly better for the \textit{JCN-S} and \textit{WuP} mappings). Following them are the \textit{Deepwalk} models, which in turn outperform \textit{node2vec} and \textit{TransR}. In the dynamic synset selection setup (see Table \ref{tab:eval_model_based}), all the models except \textit{node2vec} and \textit{TransR} are superior to raw WordNet, and the best models are \textit{FSE} and \textit{path2vec ShP/WuP}, significantly outperforming the others. \textit{Path2vec} models trained on \textit{JCN-S} and \textit{LCH} are on par with \textit{Deepwalk} and much better than \textit{node2vec} and \textit{TransR}. 
We believe it to interesting, considering that it does not use random walks on graphs and is conceptually simpler than \textit{FSE}.

Note that word embedding models trained on text perform \textit{worse} than the WordNet-based embeddings (including \textit{path2vec}) on the semantic similarity task. For example, the \textit{word2vec} model of vector size 300 trained on the Google News corpus \cite{Mikolov:2013} achieves Spearman correlation of only 0.449 with SimLex999, when testing only on nouns. The \textit{GloVe} embeddings \cite{pennington2014glove} of the same vector size trained on the Common Crawl corpus achieve 0.404. 

\begin{table}
\footnotesize
\centering
\begin{tabular}{lc}
\toprule
\textit{Model} & \textit{Correlation} \\

\midrule
TransR~\cite{transr:2015} & 0.386  \\ 
node2vec~\cite{grover2016node2vec} & 0.462  \\
Deepwalk~\cite{perozzi2014deepwalk} & 0.533  \\
FSE~\cite{subercaze:2015} & \textbf{0.556}  \\
\midrule
Raw WordNet JCN-S & 0.487  \\
Raw WordNet LCH & 0.513  \\
Raw WordNet ShP & 0.513  \\
Raw WordNet WuP & 0.474  \\
\midrule

\textit{path2vec} JCN-S & 0.533  \\
\textit{path2vec} LCH & 0.532  \\
\textit{path2vec} ShP & \textbf{0.555}  \\
\textit{path2vec} WuP & \textbf{0.555}  \\
\bottomrule
\end{tabular}
\caption {Spearman correlations with human SimLex999 noun similarities (model synset selection).}
\label{tab:eval_model_based}
\end{table}

\section{Experiment 2: Extrinsic Evaluation based on Word Sense Disambiguation}
\label{subsec:wsd}

\paragraph{Experimental Setting}

As an additional extrinsic evaluation, we turned to word sense disambiguation task, reproducing the WSD approach from \cite{sinha2007unsupervised}. The original algorithm uses WordNet similarities; we tested how using dot products and the learned embeddings instead will influence the WSD performance.

\begin{figure*}
       \centering
       \includegraphics[width=0.7\textwidth]{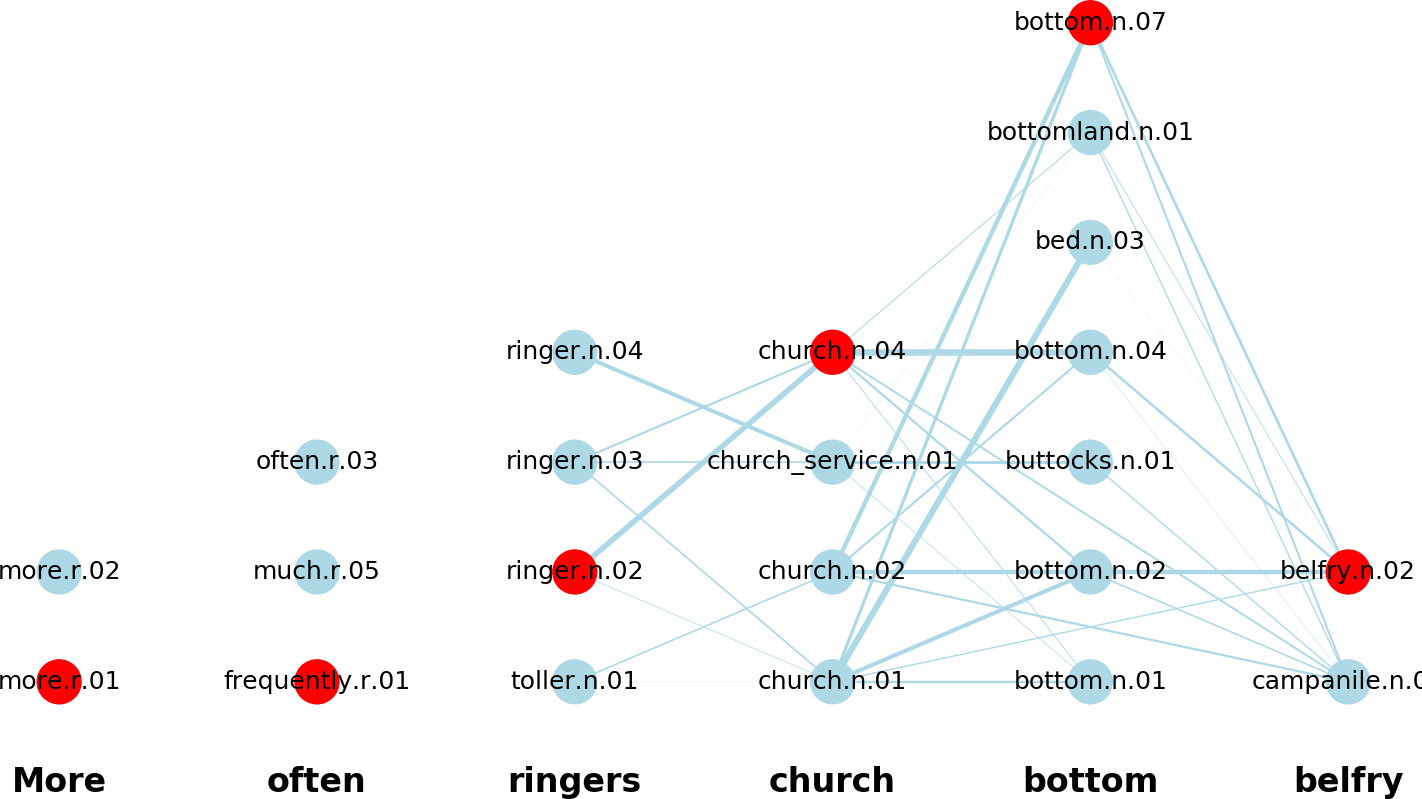}
       \caption{A sentence graph for WSD, where a column lists all the possible synsets of a corresponding word.}
       \label{fig:graph_wsd_example}
\end{figure*}

The employed  WSD algorithm starts with building a graph where the nodes are the WordNet synsets of the words in the input sentence. The nodes are then connected by edges weighted with the similarity values between the synset pairs (only if the similarity exceeds a threshold, which is a hyperparameter; we set it to 0.95). The final step is selecting the most likely sense for each word based on the weighted in-degree centrality score for each synset (in case of ties, the first synset is chosen). Figure \ref{fig:graph_wsd_example} shows a  graph generated for the sentence `\textit{More often than not, ringers think of the church as something stuck on the bottom of the belfry}'. Note that we disambiguate nouns only. 

\paragraph{Discussion of Results}

Table \ref{tab:eval_wsd} presents the WSD micro-F1 scores using raw WordNet similarities, 300D \textit{path2vec}, \textit{Deepwalk} and \textit{node2vec} models, and the 128D \textit{FSE} model. We evaluate on the following all-words English WSD test sets: Senseval-2 \cite{senseval:2001}, Senseval-3 \cite{senseval3:2004}, and SemEval-15 Task 13 \cite{semeval2015-2}. Raw WordNet similarities are still the best, but the \textit{path2vec} models are consistently the second after them (and orders of magnitude faster), outperforming other graph embedding baselines. The largest drop between the original and vector-based measures is for \textit{JCN-S}, which is also the only one which relies not only on graph but also on external information from a corpus, making it more difficult to approximate (see also Figure \ref{fig:p_histo}, where this measure distribution seems to be the most difficult to reproduce). Note that both the original graph-based measures and graph embeddings do not outperform the most frequent sense (MFS) baseline, which is in line with the original algorithm~\cite{sinha2007unsupervised}.

Here our aim was not to improve WSD systems but to compare \textit{path2vec} against other graph embedding methods in an extrinsic, task-based evaluation. This is also the reason why we do not compare against many other existing WordNet-based WSD systems: we are interested only in the approaches which learn dense representations of graph nodes, as \textit{path2vec} does.

\begin{table}
\footnotesize
\centering
\begin{tabular}{lccc}
\toprule
Model & Senseval2 & Senseval3 & SemEval-15 \\
\midrule
Random sense & 0.381  & 0.312 & 0.393 \\
\midrule

\multicolumn{4}{c}{ \textit{Baselines}  (various graph embeddings)} \\
\midrule
TransR  & 0.540  & 0.466 & 0.536  \\
node2vec  & 0.503 & 0.467 & 0.489  \\
Deepwalk  & 0.528 & 0.476 & 0.552  \\
FSE  & 0.536 & 0.476 & 0.523 \\

\midrule

\multicolumn{4}{c}{ \textit{WordNet} (graph-based measures)} \\
\midrule
JCN-S & \textbf{0.620} & \textbf{0.558} & \textbf{0.597} \\
LCH & 0.547 & 0.494 & 0.550  \\
ShP & 0.548 & 0.495 & 0.550  \\
WuP & 0.547 & 0.487 &  0.542 \\

\midrule
\multicolumn{4}{c}{ \textit{path2vec} (vector-based measures)} \\
\midrule
JCN-S & 0.511 & 0.463 & 0.508  \\
LCH & 0.527 & 0.472 & 0.536  \\
ShP & 0.534 & \textbf{0.489} & \textbf{0.563}  \\
WuP & \textbf{0.543} & \textbf{0.489} & 0.545 \\
\bottomrule
\end{tabular}
\caption {F1 scores on all-words WSD tasks.}
\label{tab:eval_wsd}
\end{table}

\section{Computational Efficiency Evaluation}
\paragraph{Pairwise Similarity Computation}

One of the reasons to use \textit{path2vec} embeddings is computational efficiency. Directly employing the WordNet graph to find semantic similarities between synsets is expensive. The dot product computation is much faster as compared to shortest path computation (and other complex walks) on a large graph. Also, dense low-dimensional vector representations of nodes take much less space than the pairwise similarities between all the nodes. 

 The time complexity of calculating the shortest path between graph nodes (as in \textit{ShP} or \textit{LCH}) is in the best case linear in the number of nodes and edges \cite{leacock1998combining}. \textit{JCN-S} compares favorably since it is linear in the height of the taxonomy tree \cite{jiang1997}; however, it still cannot leverage highly-optimized routines and hardware capabilities, which makes the use of vectorized representations so efficient. Calculating Hamming distance between binary strings (as in the \textit{FSE} algorithm) is linear in the sum of string lengths, which are equivalent of vector sizes \cite{hamming1950error}.  At the same time, the complexity of calculating dot product between float vectors (as in \textit{path2vec}) is linear in the vector size by the definition of the dot product and is easily and routinely parallelized.
 
 As an example, let us consider the popular problem of ranking the graph nodes by their similarity to one particular node of interest (finding the `nearest neighbors').  Table \ref{tab:speedup} shows the time for  computing similarities of one node to all other WordNet noun nodes, using either standard graph similarity functions from NLTK, Hamming distance between 128D binary embeddings, or dot product between a 300D float vector (representing this node) and all rows of a $82115 \times 300$ matrix. Using float vectors is 4 orders of magnitude faster than  \textit{LCH},  3 orders faster than \textit{JCN}, and 2 orders faster than Hamming distance. 

\begin{table}
\footnotesize
\centering
\begin{tabular}{lc}
\toprule
\textit{Model} & \textit{Running time} \\
\midrule
\textit{LCH} in NLTK & 30 sec. \\
\textit{JCN-S} in NLTK & 6.7 sec. \\
FSE embeddings  & 0.713 sec. \\
\textit{path2vec} and other float vectors & \textbf{0.007} sec. \\
\bottomrule
\end{tabular}
\caption {Computation of 82,115 similarities between one noun synset and all other noun synsets in WordNet.}
\label{tab:speedup}
\end{table}

\paragraph{Construction of the Training Set} 
Despite its computational efficiency at test time, constructing a training dataset for \textit{path2vec} (following the workflow described in Section \ref{sec:datasets}) requires calculating pairwise similarities between all possible pairs of graph nodes. This leads to a number of similarity calculations quadratic in the number of nodes, which can be prohibitive in case of very large graphs. However, instead of this, the training datasets for \textit{path2vec} can be constructed much faster by taking the graph structure into account. In essence, this implies finding for each node $v$ the set of other nodes directly connected to it or to its direct graph neighbors (set of second order graph neighbors, $V_2$).  Then, graph similarity is calculated only for the pairs consisting of each $v$ and the nodes in their respective $V_2$; these pairs constitute the training dataset (the same thresholds and normalization procedures apply). 

The amount of pairwise similarity calculations is then linear in the number of nodes times the average number of neighbors in $V_2$, which is much better. Particularly, in the case of WordNet, each node (synset) has 36 synsets in its $V_2$ on average, and half of the nodes do not have any neighbors at all.
Thus, only 2,935,829 pairwise similarity calculations are needed, 1,000 times less than when calculating similarities between all synset pairs. 

Following that, e.g., the training dataset for \textit{JCN-S} can be constructed in 3 minutes, instead of 5 hours, with similar speedups for other graph distance measures. The training datasets constructed in this `fast' way showed negligible performance decrease compared to the `full' datasets ($0.07...0.03$ drop in the semantic similarity experiments, and  $<0.03$ drop in the WSD experiments). It means that when using \textit{path2vec} in practical tasks, one can construct the training dataset very quickly, preserving embeddings performance.

\section{Discussion and Conclusion}
We presented \textit{path2vec}, a  simple, effective, and efficient model for embedding graph similarity measures. It can be used to learn vector representations of graph nodes, approximating shortest path distances or other node similarity measures of interest. Additionally, if the similarity function is based on the shortest path, this paves the way to a quick and efficient calculation of the shortest distance between two nodes in large graphs.

Our model allow for much more efficient graph distances calculations (3 or 4 orders of magnitude faster depending on a similarity measure). In applications one could replace path-based measures with dot product between \textit{path2vec} embeddings, gaining significant speedup in distance computation between nodes. Thus, our model could be used to speed up various other graph-based algorithms that make use of node distance computations, such as \citet{floyd1962algorithm} algorithm, \citet{dijkstra1959note} algorithm, or algorithms for computing node betweenness centrality~\cite{brandes2001faster}.

In this paper, we used our model to learn embeddings of WordNet synsets and showed that in the semantic similarity task, the resulting representations perform better than the state-of-the-art graph embedding approaches based on random walks. Interestingly, the learned embeddings can  outperform the original WordNet similarities on which they were trained. \textit{path2vec} was also evaluated on the WSD task (it has not been done before for graph embeddings, to our knowledge), again outperforming other approaches.

However, \textit{path2vec} can be trained on arbitrary graph measures and is not restricted to the shortest path or to only tree-structured graphs. In the future, we plan to explore the possibility of training embeddings able to approximate multiple similarity metrics at once. Another direction of further research is to apply our model to other types of data, such as social networks or graph of roads.

\section*{Acknowledgements}
This work has been partially supported by Deutsche Forschungsgemeinschaft (DFG) within the JOIN-T (grant BI 1544/4-1 and SP 1999/1-1)) project and the ACQuA project (grant BI 1544/7-1 and HA 5851/2-1), which is part of the Priority Program “Robust Argumentation Machines (RATIO)” (SPP-1999), and Young Scientist Mobility Grant from the Faculty of  Mathematics and Natural Sciences, University of Oslo. We thank three anonymous reviewers for their most useful feedback. Last but not least, we are grateful to Sarah Kohail who helped with computing the first version of the \textit{node2vec} baselines.

\bibliography{path2vec}
\bibliographystyle{acl_natbib}

\end{document}